# Densely Dilated Spatial Pooling Convolutional Network using benign loss functions for imbalanced volumetric prostate segmentation

Qiuhua Liu #, Min Fu#, Hao Jiang, Xinqi Gong*

janeliu@ruc.edu.cn, fm20080525@ruc.edu.cn, jiangh@ruc.edu.cn, xinqigong@ruc.edu.cn
#Both are first authors.
*Corresponding to: Xinqi Gong, xinqigong@ruc.edu.cn

## Abstract

The high incidence rate of prostate disease poses a requirement in early detection for diagnosis. As one of the main imaging methods used for prostate cancer detection, Magnetic Resonance Imaging (MRI) has wide range of appearance and imbalance problems, making automated prostate segmentation fundamental but challenging. Here we propose a novel Densely Dilated Spatial Pooling Convolutional Network (DDSP ConNet) in encoder-decoder structure. It employs dense structure to combine dilated convolution and global pooling, thus supplies coarse segmentation results from encoder and decoder subnet and preserves more contextual information. To obtain richer hierarchical feature maps, residual long connection is furtherly adopted to fuse contexture features. Meanwhile, we adopt DSC loss and Jaccard loss functions to train our DDSP ConNet. We surprisingly found and proved that, in contrast to re-weighted cross entropy, DSC loss and Jaccard loss have a lot of benign properties in theory, including symmetry, continuity and differentiability about the parameters of network. Extensive experiments on the MICCAI PROMISE12 challenge dataset have been done to corroborate the effectiveness of our DDSP ConNet with DSC loss and Jaccard loss. Totally, our method achieves a score of 85.78 in the test dataset, outperforming most of other competitors.

## 1 Introduction

Prostate disease, including prostate cancer, prostatitis and enlarged prostate, occurs usually on older men[1]. Especially, in the United States, approximately 11.6% of men will be diagnosed with prostate cancer during his lifetime [2]. Meanwhile, as the wider spread of computer-aided diagnosis, accurate segmentation of prostate in T2-weighted Magnetic Resonance Imaging (MRI) is fundamental and highly demanded.

So far, however, precisely automated prostate segmentation from MRI is still a challenging task for several reasons [Yu

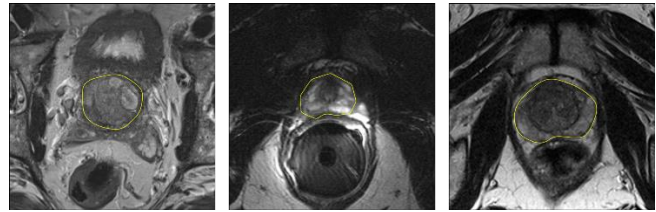

Figure 1. Examples of prostate in MRI in the dataset of PROMISE12, only showing the center slices, where the yellow contours indicates segmented ground truth.

L et al, 2017]. Firstly, due to variation of intensity distribution and field homogeneity, wide range of prostate shape in different scans brings high difficulties to prostate segmentation. Secondly, the variation in the scanners leads to the global inter-scan variability and intra-scan intensity variation. Thirdly, the lack of clear boundary between prostate and its surrounding makes accurate segmentation even harder.

Over the past years, in the task of prostate segmentation, many automated methods have been proposed. A few methods used atlases [Klein S et al, 2007, Martin S et al, 2010] and the segmented masks were registered in image similarity metrics, which relied on the similarity between the atlas. Based on sound theory from mathematics, some [Toth R et al, 2012] methods adopted deformable models including active shape models and level sets, while weak edge information could lead to poor performance. Recently, as the uncovering of deep neural network's high capability in extracting features, more deep learning techniques have been utilized in prostate segmentation tasks. For example, V-net [Milletari F et al, 2016] produced an end-to-end approach for 3D medical image segmentation based on a volumetric fully convolutional neural network.

Recognizing the drawbacks of previous methods at prostate segmentation and the great ability of CNN in learning richer features, we designed a new Densely Dilated Spatial Pooling Convolutional Network (DDSP ConNet) in encoder-decoder structure. Drawing from the idea of ASPP [Chen L C et al, 2017] (Atrous Spatial Pyramid Pooling) and U-net [Ronneberger O et al, 2015], our design characterizes densely dilated spatial pooling convolutional block with high ability in extracting information in different scales. Additionally, we adopted skip connection and deconvolution operation in decode and encode subnet, ensuring the combination of features from diverse levels and remaining the size of output images. Achieved 86.42% in the dataset of PROMISE12[3], our DDSP ConNet is efficient and accurate, and surpasses most of the previous networks listed in the challenge. Figure 1 shows some examples of prostate in MRI and their segmented ground-truth in the dataset of PROMISE12.

---

[1] https://medlineplus.gov/prostatediseases.html

[2] https://seer.cancer.gov/statfacts/html/prost.html

When segmenting the prostate, if assigning the target organ as foreground while the other regions as background, we would easily find the severe data imbalance, a common problem in medical images, with the negative region far larger than the positive one. This unbalanced problem could lead to high-precision, low-recall segmentation results, because the learning process may converge to local minima of a sub-optimal loss function, thus outcomes may strongly bias towards non-target ones. In computer-aided diagnosis, however, due to the great significance of high recall in automated segmented system, this result is certainly undesired. Facing this data imbalance problem, a common solution was down-sampling [Valverde S and Cabezas M et al, 2017], but it could miss some information contented in the images and the segmented results may bias towards rare classes. Additionally, some loss functions [Pereira S and Pinto A et al, 2016, Salehi S S M et al, 2017] based on sample re-weighting were designed for this imbalance problem. For instance, organ or lesion region was given more importance during training [Salehi S S M et al, 2017]. However, they did not perform well in highly unbalanced data.

Aiming to solve the imbalance problem and overcome some shortcomings of traditional cross entropy loss function, in training neural network, such as overstressing the difference between each voxel, asymmetry and the cost easily going to infinity, we employed DSC loss from Dice Similarity Coefficient and Jaccard loss from Jaccard Similarity Coefficient to train our network. In this paper, apart from simply adopting the Dice Similarity Coefficient [Milletari F et al, 2016] and Jaccard Similarity Coefficient, we also found some benign properties of them and showed their great benefits in imbalanced data through experiments. Formally, contrast to the cross entropy overstressing the difference of each voxel between predicted outputs and ground-truth images, these two loss functions consider the similarity of the whole mask. Additionally, we theoretically proved that superior to the cross entropy, both of them are continuous and differentiable in parameters of network almost everywhere. Furtherly, we compared the topology strength of these loss functions and found DSC and Jaccard loss much more sensitive in the proposition that [Martin Arjovsky et al, 2017] the weaker the distance, the easier it is to define a continuous mapping from parameter space to output mask space and to reach a converged network.

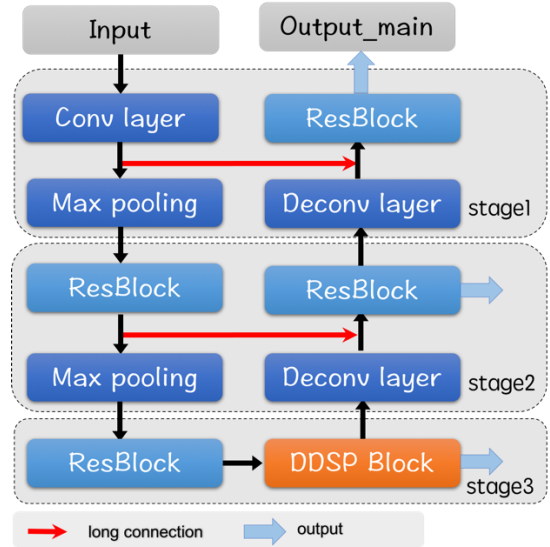

Figure2. The whole architecture of our DDSP ConNet containing three stages sharing same resolutions. It is based on encoder-decoder structure with DDSP block and residual long connection (shown in orange arrowhead).

Experiments have been done to verify their capability in dealing with imbalanced data and good properties.

The contributions of our paper are:

1. We accurately segmented imbalanced prostate MRI data using a unique network with appropriate loss functions.

2. We proposed a new network architecture (DDSP ConNet), which could highly integrate multi-scale context and furtherly improve the efficiency of the whole network.

3. We found and proved the benign properties of DSC loss and Jaccard loss functions for imbalanced medical image data.

## 2   Method

In this section, we will describe the formulation of DDSP ConNet for accurate prostate segmentation and its whole structure can be seen in Figure 2. We start by introducing the whole encoder-decoder structure for end-to-end training. Furthermore, to obtain richer hierarchical feature maps, residual long connection was adopted to fuse contexture features between encoder and decoder, and multi-supervision structures were utilized to generate better prostate results. Then we elaborate our novel DDSP block for accurate prostate segmentation, which densely connects high level contexture features and image level ones.

### 2.1   Encoder-decoder structure with multi-level contextual features

Our DDSP ConNet is based on encoder-decoder structure, like most CNN models for semantic segmentation [Ronneberger O et al, 2015]. The encoder, also called an up-to-down path and made of convolution and pooling layers, is responsible for extracting the feature maps of input images, enlarging receptive fields and down-sampling the extracted feature maps. Relatively, consisting of deconvolution and convolution layers and functioning on up-sampling the feature maps, the decode subnet is a down-to-up path, which outputs voxel-to-voxel segmentation predictions. For convenience, we divide the whole encoder-decoder structure into three stages. As shown in Figure 2, the feature maps in each stage share same resolutions.

---

[3] https://promise12.grand-challenge.org/

However, in encoder and decoder, we could only obtain some coarse segmentation results, not sufficient for accurate prostate segmentation. To share features of local information from encoder with decoder and obtain hierarchical feature maps, residual long connection is adopted to fuse contexture features between them. Different from concatenate long connection proposed in U-net, residual long connection proposed in CUMED [Yu L et al, 2017] ensures more precise localization of segmentation targets and performs better.

Additionally, noticing that training such a deep ConNet can easily get trouble in gradient vanishing, we adopted multi-supervision structure to strengthen the training process. Not only could this auxiliary structure effectively encourage the gradient in back-propagation to flow by calculating auxiliary loss of middle layers in encoder, but also acquire multi-level contexture features. In the test stage, accordingly, we fused the multi-level contextual features to get better segmentation predictions. We call this operation as multi-fusion, thus the loss in our network is:

$$\text{LOSS} = \alpha L_{main} + \beta L_{stage2} + \gamma L_{stage3}$$

Here, $\alpha, \beta, \gamma$ are all weights in our network. $L_{main}$, $L_{stage2}$ and $L_{stage3}$ are loss values of main output, stage2 output and stage3 output respectively.

## 2.2 Densely Dilated Spatial Pooling block

In a deep neural network, the size of receptive field roughly indicates the amount of context information we use [H. Zhao et al, 2017]. To exponentially enlarge it, most ConNet utilized pooling layers or convolution strides, but not sufficient to obtain global features. Inspired by [H. Zhao et al, 2017], we adopted a global pooling layer to extract image level features highly beneficial to localize the target object.

However, the pooling operation makes ConNet model invariant to local image transform, and downssizes the pictures as well as losing voxel-wisely resolution. Therefore, we adopted dilated convolution, also called atrous convolution, to replace the pooling operation [Chen L C et al, 2017, Yu F and Koltun V, 2015]. The dilated convolution enlarges the receptive field while maintaining the resolution of responsive feature maps and effectively preserving the local information.

To acquire more accurate predictions, we tried to fuse the image level features from global pooling layers and meticulous features in dilated convolution layers. Contrast to Atrous Spatial Pyramid Pooling structure (ASPP) [Chen L C et al, 2017] concatenating the layers parallel, we adopted

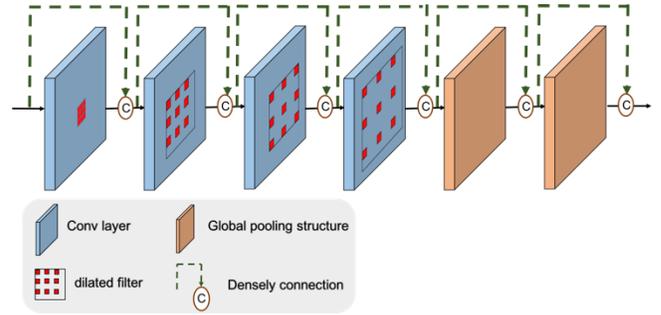

Figure3. The structure of our DDSP block contains layers in different dilated and global pooling rates, which are all connected by element-wise concatenation.

densely connection structure [Huang G et al, 2016] to deal with the multi-level features, and named this densely dilated spatial pooling block (DDSP) [Figure 3]. Not only could it concatenate the multi-level features side by side, but also highly reuse those dense features and preserve more hierarchical contexture.

## 3 Loss function (Dice loss and Jaccard loss)

When segmenting the prostate, we found that, if assigning the target organ as foreground while the other regions as background, the background is far larger than foreground. Facing this data imbalance problem, we tried to use DSC loss and Jaccard loss based on the Dice Similarity Coefficient and Jaccard Similarity Coefficient to train our DDSP ConNet. The definition of DSC Loss:

$$L_{DSC}(P_r(x), P_\theta(x)) = 1 - \frac{2P_\theta^T P_r}{\|P_\theta\|_2^2 + \|P_r\|_2^2}$$

Here, for a brief statement, $P_\theta$ and $P_r$ are two multi-dimensional distributions, and $P_r \in X^N, P_\theta \in X^N$. $X \in [0,1]$. $\|\cdot\|_2$ denotes the l2 norm. Similarly, based on the Jaccard Similarity Coefficient, we adopt the Jaccard loss.

The definition of Jaccard loss:

$$L_{Jaccard}(P_r(x), P_\theta(x)) = 1 - \frac{P_\theta^T P_r}{\|P_\theta\|_2^2 + \|P_r\|_2^2 - P_\theta^T P_r}$$

Denotes here are all the same with Dice loss.

Apart from simply adopting these loss functions into the data imbalance problem, we have explored many benign properties of them, and we will further analyze them in theory.

To show the properties of loss functions more clearly, we also give some analysis of KL divergence. For example, it could easily go to infinite and could not provide reasonable gradient, and it is not symmetrical. The detail of analysis could be found in the Appendix A.

### 3.1 Differentiability of Dice loss and Jaccard loss

Different from the KL divergence, these two loss functions have more benign properties in differentiability. The differentiability could make the network more stable and prevent the parameters from being trapped in the local optimum.

**Theorem two**. Let $P_r$ be a fixed point over $X^N$. Let z be a random variable over other space Z. Let g: $Z \times R^d \to X^N$ be a function, that will be denoted $g_\theta(z)$ with z the first coordinate and θ the second. Let $P_\theta$ denotes the outcome of $g_\theta(z)$, and $P_\theta \in X^N$. Then,
1. If g is continuous in θ, so is $L_{DSC}(P_r, P_\theta)$.
2. If g is locally Lipchitz and satisfies regularity assumption 1, then $L_{DSC}(P_r, P_\theta)$ is continuous and differentiable almost everywhere.

Proof: See Appendix B.

The following corollary shows that learning by minimizing the DSC loss makes sense (at least at theory) with neural networks in the task of semantic segmentation.

**Corollary one**. Let $g_\theta$ be any feedforward neural network[4] parameterized by θ. Then assumption one is satisfied and therefore $L_{DSC}(P_r, P_\theta)$ is continuous everywhere and differentiable almost everywhere.

Proof: See Appendix C.

Additionally, statement 1 and 2 are both true to Jaccard loss, thus Jaccard loss is also a sensitive loss function. Detailed explanation could be seen in Appendix D.

### 3.2 The topology strength of loss functions

The topology strength of loss function means the complexity of optimization when training the neural network. That the topology induced by ρ is weaker than that of ρ' means that the set of convergent sequences under ρ is a superset of that under ρ' [Martin Arjovsky et al, 2017]. A weaker loss, indicates that it is easier to find an appropriate set of parameters, for it is easier for the output distributions to converge into the target distributions.

Before comparing the relative topology strength of these loss functions, we draw into another distance for our proof.

The Total Variation (TV) distance:
$$\delta(P_r, P_\theta) = sup_i |P_r - P_\theta|$$

Here, $P_r$ and $P_\theta$ denotes two multi-dimension distributions, $P_r \in X^N, P_\theta \in X^N$, and i denotes the index of dimension in distribution.

The following theorem describes the relative strength of the topologies induced by these distance and divergences, with KL the strongest, followed by TV, and DSC loss and Jaccard loss.

**Theorem three**. Let P be a point and $P \in X^N$. $(P_n)_{n \in N}$ be a sequence on $X^N$. Then considering all limits as $n \to \infty$,
1、 $L_{Jaccard}(P_n, P) \to 0$ with the Jaccard loss.
2、 $L_{DSC}(P_n, P) \to 0$ with the DSC loss implies the statements in 1.
3、 $\delta(P_n, P) \to 0$ with δ the total variation distance implies the statements in 2.
4、 $KL(P_n||P) \to 0$ or $KL(P||P_n) \to 0$ imply the statement in 3.

Proof: See Appendix E.

Theorem three shows that KL and TV distances are not sensible enough, and DSC loss and Jaccard loss are more ideal loss functions.

### 3.3 DSC loss layer and Jaccard loss layer

For convenience, here we set $\sum_{i=0}^{n^3} P_\theta(x)_i^2 + \sum_{i=0}^{n^3} P_r(x)_i^2$ as U, $\sum_{i=0}^{n^3} P_\theta(x)_i P_r(x)_i$ as K, $\sum_{i=0}^{n^3} P_\theta(x)_i + \sum_{i=0}^{n^3} P_r(x)_i$ as L, and $\frac{P_\theta(x)^T P_r(x)}{\|P_\theta(x)\|_2^2 + \|P_r(x)\|_2^2}$ as $\overline{DSC}$.

In the training process, $P_r$ denotes the ground − truth segmented mask, $P_\theta$ denotes the output, and $X^N$ denotes the space of segmented mask. N is the sum of voxel.

We have DSC Loss layer:

$$L_{DSC} = 1 - \frac{2\sum_{i=0}^{n^3} P_\theta(x)_i P_r(x)_i}{\sum_{i=0}^{n^3} P_\theta(x)_i^2 + \sum_{i=0}^{n^3} P_r(x)_i^2} = 1 - \frac{2P_\theta(x)^T P_r(x)}{\|P_\theta(x)\|_2^2 + \|P_r(x)\|_2^2}$$

For a brief statement, we could see the input and output images as n*n*n volumes[5], and the dimension of vector is $n^3$. *x* denotes the input image, $P_\theta(x)$ denotes the output image, and the $P_r(x)$ denotes the ground-truth image. *i* is the index of the vectors, θ denotes all the parameters of the whole network and $P_\theta(x)_i$ or $P_r(x)_i$ denotes a voxel. The value of $P_r(x)_i$ could only be 0 (background, non-prostate) or 1 (foreground, prostate) and that of $P_\theta(x)_i$ ranges from 0 to 1.

The formulation of Dice loss can be differentiated yielding the gradient:

$$\frac{\partial L_{DSC}}{\partial P_\theta(x)_i} = -2 * \frac{P_r(x)_i * U - 2K * P_\theta(x)_i}{U^2}$$

To provide reasonable gradient, in contrast to the definition of Dice Similarity Coefficient, we adopt square, U, in the denominator. Here we give an example for further explanation.

**Example** If we do not employ square in the denominator in the DSC loss, that is

$$L_{DSC} = 1 - \frac{2K}{L} = 1 - \frac{2P_\theta(x)^T P_r(x)}{\|P_r(x)\| + \|P_\theta(x)\|}$$

The gradient is

$$\frac{\partial L_{DSC}}{\partial P_\theta(x)_i} = -2 * \frac{P_r(x)_i * L - 2K}{L^2}$$

Under this definition, sharing the same ground-truth and prediction image, the voxels belonging to foreground in the ground-truth, have the same gradient whatever their prediction values, thus DSC loss without square in denominator could not provide reasonable gradient when training the network.

---

[4] By a feedforward neural network we mean a function composed by affine transformations and pointwise nonlinearities which are smooth Lipschitz functions (such as the sigmoid, tanh, elu, softplus, etc)

[5] The definition of other kind of volume such as n*n*m is similarity.

Here, interestingly, as the definition of DSC loss layer, we could easily find that only when $P_r(x)_i \neq 0$, $P_\theta(x)_i P_r(x)_i \neq 0$, which means that the DSC loss only cares about the voxel whose ground-truth belongs to prostate. To some extent, this feature explains why the DSC loss could deal with the imbalance problem.

Additionally,

when $P_r(x)_i = 0$, $\frac{\partial L_{DSC}}{\partial P_\theta(x)_i} = 4P_\theta(x)_i * \overline{DSC} * \frac{1}{U}$.

when $P_r(x)_i = 1$, $\frac{\partial L_{DSC}}{\partial P_\theta(x)_i} = \frac{-2}{U} + 4P_\theta(x)_i * \overline{DSC} * \frac{1}{U}$.

As the beginning of training process, the DSC loss was comparably high, that is a small $\overline{DSC}$, when the DSC loss layer only gives effective gradient to the prostate voxel, which is highly beneficial for network training in imbalance data.

Similarity to DSC loss layer, we have Jaccard loss layer:

$L_{Jaccard} = 1 - \frac{K}{U-K} = 1 - \frac{P_\theta(x)^T P_r(x)}{\|P_\theta(x)\|_2^2 + \|P_r(x)\|_2^2 - P_\theta(x)^T P_r(x)}$;

Denotes here are all the same with Dice loss. The formulation of Jaccard loss can be differentiated yielding the gradient:

$\frac{\partial L_{Jaccard}}{\partial P_\theta(x)_i} = -\frac{P_r(x)_i * (U-K) - K * (2P_\theta(x)_i - P_r(x)_i)}{(U-K)^2}$

computed with respect to the i-th voxel of the prediction.

## 4 Experiments and Results

### 4.1 Dataset and Pre-processing

To illustrate the effectiveness of our DDSP ConNet and loss functions, we carried extensive experiments on MICCAI Prostate MR Image Segmentation (PROMISE12) challenge dataset [Litjens G et al, 2014]. The training dataset contains 50 volumetric transversal T2-weighted MRIs of prostate and their ground-truth. For independent evaluation, the testing dataset contains 30 MRIs and their ground-truth are held by the organizer. These MRIs were collected in different hospitals, equipment and acquisition protocols, so there are variations in voxel size, dynamic range, position, field of view and anatomic appearance [Yu L et al, 2017]. During the pre-processing step, similar to V-net [5], firstly we used the N4 bias filed correction function of the ANTs framework to normalize the datasets and then resampled them to a common resolution of 1*1*1.5 mm. By varying the position of the control points with random quantities obtained from Gaussian distribution with zero mean and 15 voxels standard deviation, we applied random deformations to the training scans. Furthermore, we centrally cropped the 3D MRI data to size of 96*96*48 pixels.

### 4.2 Training

**Initialization** We implemented our model in Caffe and trained the model from sketchy with one Nvidia TITAN Xp GPU. The convolution layers were initialized with bias b=0, a Gaussian weight filled with Standard Deviation=0.01, and a bilinear 3D weight. Each layer consists of batch norm and ReLU activation function.

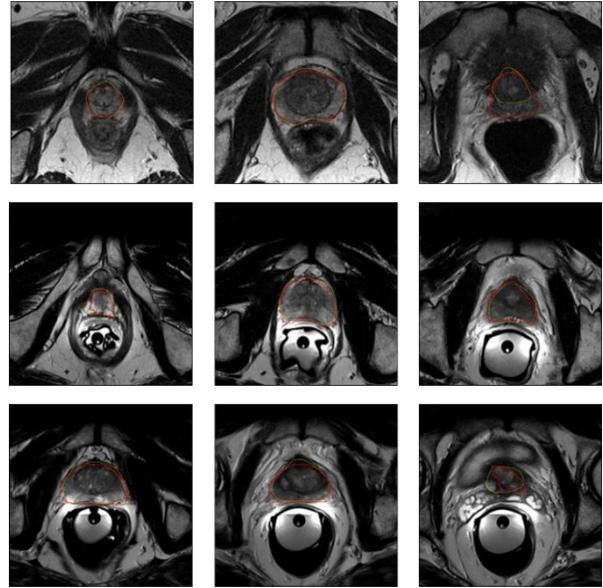

Figure 4: Examples of qualitative segmentation results. Each row from top to down are case 16, case 22 and case 28 at the apex(left), center (middle) and base (right) of the T2-weighted transverse prostate in testing dataset. Ground-truth are shown in yellow contour, and segmentation results shown in red one. All these results are directly obtained from the PROMISE12 challenge website.

**Optimization** We adopted stochastic gradient descent (SGD) to train our network. Due to the limit of the GPU, we set batch size in two. Besides, we used an initial learning rate of 0.001, weight decay of 0.005 and momentum of 0.99, and the learning rate reduced 80% in every 10000 iterations. Training time of our model ranged between 8 and 10 hours.

### 4.3 Comparison to State of the Art

There are several kinds of evaluation metrics used in the PROMISE12 challenge, including the DSC (Dice Similarity Coefficient), aRVD (the percentage of the absolute difference between the volumes), ABD (the average over the shortest distance between the boundary points of the volumes) and 95 HD (95% Hausdorff distance) [Yu L et al, 2017]. In these evaluations, higher DSC values means better performance of model while lower scores of others point to better performance, and these evaluations make up the final score. In our testing stage, all of them were used to evaluate our model but with only DSC in training stage. Figure 4 shows some qualitative results and all of our detailed results could be found in the website[6] of the PROMISE12 challenge.
Table1 below illustrates the performance of our DDSP ConNet (Named RUCIMS in the web) in the testing set of PROMISE12 and that of some other competitors. Outperforming most of other methods, our approach achieved score of 85.78, which shows the high capability of our approach in extracting and preserving information as well as strong robust.

Table1: The performance of different models in the training of PROMISE12.

| Team Name | Score |
|---|---|
| CUMED | 86.65 |
| **RUCIMS (ours)** | 85.78 |
| CREATIS | 85.74 |
| methinks | 85.41 |
| **MedicalVision** | 85.33 |
| **BDSlab** | 85.16 |

### 4.4 Ablation Study for DSC loss and jaccard loss function and DDSP block

We ran extensive experiments to analyze the behavior of different loss functions and DDSP block for semantic (prostate) segmentation. For all these experiments training in different loss functions with various blocks, we utilized our basic model with residual long connection, multi-supervise, while other parameters in the network remained the same in these several experiments.

Table 2: The performance in term of dice coefficient of network training different loss functions in various blocks.

|  | reweight-ce | dsc | jaccard |
|---|---|---|---|
| non | 83.92% | 85.01% | 86.14% |
| ddsp | 86.64% | 86.66% | 86.38% |
| aspp | 84.97% | 85.71% | 86.16% |

From the table 2, we could easily find that as the loss function to train network and complete the task of prostate segmentation, DSC loss gets the DSC score of 86.66% in DDSP block, and outperforms the other two loss functions (Jaccard loss, re-weighted cross entropy), illustrating the its higher capacity in dealing with imbalanced medical image data. At the same time, Jaccard loss performs excellently with non or ASSP block, also indicating its ability in dealing with imbalance data. We could also see that, when remaining the same loss function in training network, the use of DDSP block improved the segmented performance. These experiments clearly show that the combination of densely concatenation, dilated layers and global pooling layers has strong robust and increases the ability in preserving context in diverse levels and thus improves the segmented model, obviously outperforming the non and parallel one. In conclusion, our DDSP block combined with DSC loss layer enjoys the highest scores.

### 4.5 Analysis of Dense Dilated Spatial Pooling block

We carried experiments on DDSP blocks with different dilation and global pooling rates. During these experiments, we utilized our DDSP network architecture with the DSC loss layer and multi-supervision. Additionally, all the other parameters remained the same.

Table 3. Performance of our model on Dice coefficient with different DDSP rate configurations. The combination of the densely concatenation, the dilated layers and global pooling layers is our DDSP block.

| Dilate rate | Global Pooling rate | Dice coefficient [%] |
|---|---|---|
| **[1,2,3,4]** | [2,4,6] | 86.66 |
| **[1,2,3,4]** | - | 86.01 |
| - | [2,4,6] | 85.98 |

From table 3, we could easily see that, without dilated convolutional layers, the network got the DSC score of 85.98, about 0.7 lower than that in DDSP block Besides, without global pooling layers, the network achieved the DSC score of 86.01, about 0.6 lower than that in DDSP block To sum up, the decline of DSC score in experiments without dilated layers or global layers clearly show that in the DDSP module, both the dilated layers and global pooling layers are necessary and essential when constructing a precise segmented network.

**For further illustration of our designs in different parts of our DDSP network and selection of loss functions when training network, we have Appendix F part, including comparison of convergence speed in different loss functions, ablation study for residual long connection, ablation study for multi-supervision, and ablation study for multi-fusion output.**

## 5 Conclusion

In this work, to complete the task of imbalanced prostate segmentation automatically and precisely, we designed a new Densely Dilated Spatial Pooling (DSPP) ConNet. Additionally, facing the data imbalance problem in the segmentation task, we employed DSC loss and Jaccard loss to train our network as well as exploring and proving some benign properties of them as loss functions. Our design of network and selection of loss functions have been validated by many extensive experimental analysis and they reached state-of-the-art accuracy on the challenging PROMISE12 dataset.

## Acknowledgments

This research was supported by State Key Laboratory of Membrane Biology to Xinqi Gong in Renmin University of

---

[6] https://promise12.grand-challenge.org/resultpro/?id=RUCIMS&folder=20180125000157_2717_RUCIMS

China. Dr. Jiang Hao has dedicated her time to evaluate our research and revise our paper.

## Appendix A. Cross entropy loss function and KL divergence

Firstly, let us review cross entropy, a common loss function in training network. As we all know, the cross entropy

$$H(P_r, P_\theta) = P_r(x) \log P_\theta(x) = H(P_r) + KL(P_r || P_\theta)$$

and The Kullback-Leibler (KL) divergence

$$KL(P_r||P_\theta) = \int log(\frac{P_r(x)}{P_\theta(x)})P_r(x)d\mu(x)$$

When we try to minimize the cross entropy or the re-weight one in the training process, actually we are minimizing the KL divergence. Here we would show some drawbacks of KL divergence.

Firstly we give theorem one showing that we won't be able to learn anything by backproping it, and the proof is so trivial that we just leave it to the reader.

**Theorem one.** Let $P_r$, $P_\theta$ be two distributions, and $P_r \in Z^N, P_\theta \in Z^N$: When $P_r$ and $P_\theta$ are not intersect, that is $P_r(x) \neq P_\theta(x), x = 1 \ldots \ldots N$, and $P_r \neq [0 \ldots \ldots 0]^T$, we have $KL(P_r||P_\theta)=\infty$.

The theorem one clearly shows that even two distributions are similar (i.e. Euclidean distance between them is small) but not intersect, the KL divergence goes to infinite, not to mention providing reasonable gradient. Therefore, KL divergence is not a good evaluation of the similarity, and it is not really possible to minimize it by gradient descent when training a network.

Additionally, the KL divergence is not symmetrical between $P_r$ and $P_\theta$, thus gives extremely unbalanced punishment when training network. Considering again of its definition, we could easily get that:

If $P_r(x)>P_\theta(x)$, x is a point belonging to foreground. When $P_\theta(x) \to 0$, the integrand inside the KL grows quickly to infinity, meaning that the loss function gives an extremely high cost to the foreground point assigned to background.

If $P_r(x)<P_\theta(x)$, x is a point belonging to background. When $P_\theta(x) \to 1$, the value inside the KL goes to 0, meaning that the loss function pays low cost when assigning a background point into foreground.

## Appendix B
**Theorem two.** Let $P_r$ be a fixed point over $X^N$. Let z be a random variable over other space $Z$. Let g: $Z \times R^d \to X^N$ be a function, that will be denoted $g_\theta(z)$ with z the first coordinate and $\theta$ the second. Let $P_\theta$ denotes the outcome of $g_\theta(z)$, and $P_\theta \in X^N$. Actually, Z is the space of target photos, $P_r(z)$ is the ground-truth of z, and $P_\theta(z)$ is the output of the network. Then,
1. If g is continuous in $\theta$, so is $L_{DSC}(P_r, P_\theta)$.
2. If g is locally Lipchitz and satisfies regularity assumption 1, then $L_{DSC}(P_r, P_\theta)$ is continuous and differentiable almost everywhere.

Proof:
1. If g is continuous in $\theta$, and $\theta \to \theta'$,

$$\left|L_{DSC}(P_r(x), P_\theta(x)) - L_{DSC}(P_r(x), P_{\theta'}(x))\right|$$
$$= \frac{2P_\theta(x)^T P_r(x)}{\|P_\theta(x)\|_2^2 + \|P_r(x)\|_2^2} - \frac{2P_{\theta'}(x)^T P_r(x)}{\|P_{\theta'}(x)\|_2^2 + \|P_r(x)\|_2^2}$$
$$\leq \frac{2}{\|P_r(x)\|_2^4} \left( \left|P_r(x)^T \left(P_\theta(x) P_{\theta'}(x)^T P_{\theta'}(x) - P_{\theta'}(x) P_\theta(x)^T P_\theta(x)\right)\right| + \left|(P_\theta(x)^T - P_{\theta'}(x)^T) P_r(x) P_r(x)^T P_r(x)\right| \right)$$
$$= \frac{2}{\|P_r(x)\|_2^4} (|G(x)| + |P(x)|)$$

$$|G(x)| = \left|P_r(x)^T (P_\theta(x) P_{\theta'}(x)^T P_{\theta'}(x) - P_{\theta'}(x) P_{\theta'}(x)^T P_{\theta'}(x) + P_{\theta'}(x) P_{\theta'}(x)^T P_{\theta'}(x) - P_{\theta'}(x) P_\theta(x)^T P_\theta(x))\right|$$
$$\leq \left|P_r(x)^T (P_\theta(x) - P_{\theta'}(x)) \|P_{\theta'}(x)\|^2\right| + \left|P_r(x)^T P_{\theta'}(x)(\|P_{\theta'}(x)\|^2 - \|P_\theta(x)\|^2)\right|$$
$$= |G_1(x)| + |G_2(x)|$$

For g is continuous in $\theta$, when $\theta \to \theta'$ and a fixed x, we have $\|P_\theta(x) - P_{\theta'}(x)\| \to 0$. As the definition of l2 norm, we have for any i=1,2,⋯N, $|P_\theta(x)_i - P_{\theta'}(x)_i| \to 0$.

$$|G_1(x)| = \left|\sum_{i=1}^N P_r(x)_i (P_\theta(x)_i - P_{\theta'}(x)_i)\right| \|P_{\theta'}(x)\|^2$$
$$\leq \|P_{\theta'}(x)\|^2 \sum_{i=1}^N P_r(x)_i |(P_\theta(x)_i - P_{\theta'}(x)_i)| \to 0$$

$|G_2(x)| \leq \left|P_r(x)^T P_{\theta'}(x) \|P_\theta(x) - P_{\theta'}(x)\|^2\right| \to 0,$ since for all x, $|P_r(x)^T P_{\theta'}(x)| \leq N$
Therefore, $|G(x)| \to 0$.

$$|P(x)| = \left|\sum_{i=1}^N (P_\theta(x)_i - P_{\theta'}(x)_i) P_r(x)_i\right| \|P_r(x)\|^2$$
$$\leq \sum_{i=1}^N |(P_\theta(x)_i - P_{\theta'}(x)_i)| P_r(x)_i \|P_r(x)\|^2$$
$$\leq \|P_r(x)\|^2 \sum_{i=1}^N |(P_\theta(x)_i - P_{\theta'}(x)_i)| \to 0$$

Therefore, $\left|L_{DSC}(P_r(x), P_\theta(x)) - L_{DSC}(P_r(x), P_{\theta'}(x))\right| \to 0$, and $L_{DSC}(P_r, P_\theta)$ is continuous in $\theta$.

2. If g is locally Lipchitz, first of all, we prove that $L_{DSC}(P_r, P_\theta)$ is also locally Lipchitz.

Because g is locally Lipchitz, for a given pair $(\theta, z)$, there is a constant $L(\theta, z)$ and an open set U such that $(\theta, z) \in U$, such that for every pair $(\theta', z) \in U$, we have $\|g_\theta(z) - g_{\theta'}(z)\|_1 \leq L(\theta, z)(\|\theta - \theta'\|_1)$. Here, $\|\cdot\|_1$ denotes l1 norm.

Easily, we could see that $|G_1(x)| \leq \|P_{\theta'}(x)\| \sum_i |P_\theta(x)_i - P_{\theta'}(x)_i| \leq N * \sum_i |P_\theta(x)_i - P_{\theta'}(x)_i| = N * \|P_\theta(x) - P_{\theta'}(x)\|_1 \leq N * L(\theta, z)(\|\theta - \theta'\|_1)$; we could define $L_1(\theta) = N * L(\theta, z)$, and achieve $|G_1(x)| \leq L_1(\theta) \|\theta - \theta'\|_1$.

In the same way, $|G_2(x)| \leq P_r(x)^T P_{\theta'}(x)(\|P_\theta(x) - P_{\theta'}(x)\|_2) \leq N * \|P_\theta(x) - P_{\theta'}(x)\|_1 \leq N * L(\theta, z)(\|\theta - \theta'\|_1) = L_1(\theta) \|\theta - \theta'\|_1$;
$|P(x)| \leq \|P_r(x)\|^2 \sum_i |P_\theta(x)_i - P_{\theta'}(x)_i| \leq \|P_r(x)\|^2 L(\theta, z)(\|\theta - \theta'\|_1)$, we could define $L_2(\theta) = \|P_r(x)\|^2 L(\theta, z)$, and achieve $|P(x)| \leq L_2(\theta) \|\theta - \theta'\|_1$.;

Therefore, $\left|D(P_r(x), P_\theta(x)) - D(P_r(x), P_{\theta'}(x))\right| \leq (L_1(\theta) + L_1(\theta) + L_2(\theta)) \|\theta - \theta'\|$,
we could define $L(\theta) = L_1(\theta) + L_2(\theta) + L_1(\theta)$, and $\left|L_{DSC}(P_r(x), P_\theta(x)) - L_{DSC}(P_r(x), P_{\theta'}(x))\right| \leq L(\theta) \|\theta - \theta'\|$, for all $\theta' \in U_{\theta_1} \cap U_{\theta_2}$, meaning that $L_{DSC}(P_r, P_\theta)$ is locally Lipchitz. This obviously implies that $L_{DSC}(P_r, P_\theta)$ is everywhere continuous, and by Radamacher's theorem[7] we know it has to be differentiable almost everywhere.

## Appendix C
**Corollary one.** Let $g_\theta$ be any feedforward neural network[8] parameterized by $\theta$. Then assumption one is satisfied and therefore $L_{DSC}(P_r, P_\theta)$

---

[7] If U is an open subset of $R_n$, and $f: U \to R_m$ is Lipschitz continuous, then f is differentiable almost everywhere in U ; that is, the points in U at which f is not differentiable form a set of measuring zero.
[8] By a feedforward neural network we mean a function composed by affine transformations and pointwise nonlinearities which are smooth Lipschitz functions (such as the sigmoid, tanh, elu, softplus, etc)

is continuous everywhere and differentiable almost everywhere.

Proof: since $g_\theta$ is $C^1$ as a function of $(\theta, z)$, then for any fixed $\theta$ we have $|g_\theta(x)_i - g_{\theta'}(x)_i| \leq \left(\left|\frac{\partial g_\theta(x)_i}{\partial \theta_j}\right| + \varepsilon\right)|\theta_j - \theta'_j|$. When we choose $L_{ij}(\theta) = \left(\left|\frac{\partial g_\theta(x)_i}{\partial \theta_j}\right| + \varepsilon\right)$ and $L_i(\theta) = \max_j L_{ij}(\theta)$, we have $m|g_\theta(x)_i - g_{\theta'}(x)_i| \leq L_i(\theta)\|\theta - \theta'\|_1$.

Here, m is the dimension of $\theta$, that is the number of the parameters in the network.

Furtherly, when we choose $L(\theta) = \max_i L_i(\theta)$, we have $m\|g_\theta(x) - g_{\theta'}(x)\|_1 \leq N * L(\theta)\|\theta - \theta'\|_1$, and $\|g_\theta(x) - g_{\theta'}(x)\|_1 \leq \overline{L(\theta)}\|\theta - \theta'\|_1$. Therefore, in the feedforward neural network, function g is continuous in $\theta$ and locally Lipchitz.

## Appendix D

From the definitions of the DSC loss and Jaccard loss, we could easily get the relationships of them: $L_{Jaccard}(P_r, P_\theta) = \frac{2L_{DSC}(P_r, P_\theta)}{1 + L_{DSC}(P_r, P_\theta)}$. This equation shows that when the g is continuous in $\theta$, we have $L_{DSC}(P_r, P_\theta)$ is continuous in $\theta$, due to the range of DSC loss is zero to one, $L_{Jaccard}(P_r, P_\theta)$ is bounded and also continuous in $\theta$. Furtherly, because of the differentiability of DSC loss, the Jaccard loss is also differentiable almost everywhere. Therefore, Jaccard loss is also coincident with Theorem two.

**Appendix E**
**Theorem three**. Let P be a point and $P \in X^N$. $(P_n)_{n \in N}$ be a sequence on $X^N$. Then considering all limits as $n \to \infty$,
1、 $L_{Jaccard}(P_n, P) \to 0$ with the Jaccard loss.
2、 $L_{DSC}(P_n, P) \to 0$ with the DSC loss implies the statements in 1.

3、$\delta(P_n, P) \to 0$ with $\delta$ the total variation distance implies the statements in 2.

4、$KL(P_n||P) \to 0$ or $KL(P||P_n) \to 0$ imply the statement in 3.

Proof:

1. As the definition of Jaccard loss and DSC loss, we can easily get that
$$\frac{2\sum_i P_{n_i} P_i}{\sum_i P_{n_i}^2 + \sum_i P_i^2 - 2\sum_i P_{n_i} P_i} \geq \frac{2\sum_i P_{n_i} P_i}{\sum_i P_{n_i}^2 + \sum_i P_i^2}$$

And we have

$\mathbf{L}_{Jaccard}(P_n, P) = 1 - \frac{2\sum_i P_{n_i} P_i}{\sum_i P_{n_i}^2 + \sum_i P_i^2 - 2\sum_i P_{n_i} P_i} \leq \mathbf{L}_{DSC}(P_n, P) = 1 - \frac{2\sum_i P_{n_i} P_i}{\sum_i P_{n_i}^2 + \sum_i P_i^2} \to 0$

2. When $\delta(P_n, P) \to 0$, for $\forall \varepsilon, \exists N_1$, such that for $\forall i = 1......N, \forall n > N_1, |P_{n_i} - P_i| < \varepsilon$;

As the proof of theorem one, $|\mathbf{L}_{DSC}(P_n, P)| = |\mathbf{L}_{DSC}(P_n, P) - \mathbf{L}_{DSC}(P, P)| \leq \frac{2}{\|P(x)\|^4}(|G_1(x)| + |G_2(x)| + |K(x)|)$;

$$|G_1(x)| = \|P\|^2 \left|\sum_i (P_{n_i} - P_i)P_i\right| \leq \|P\|^2 \sum_i |(P_{n_i} - P_i)| < \|P\|^2 N\varepsilon \to 0$$
$$|G_2(x)| = \|P\|^2(\|P\|^2 - \|P_n\|^2) \leq \|P\|^2 \|P - P_n\|^2 < \|P\|^2 \sqrt{N}\varepsilon \to 0$$
$$|K(x)| = \|P\|^2 \left|\sum_i (P_{n_i} - P_i)P_i\right| \leq \|P\|^2 \sum_i |(P_{n_i} - P_i)| < \|P\|^2 N\varepsilon \to 0$$

Therefore, $|\mathbf{L}_{DSC}(P_n, P)| \to 0$.

3. This is an application of Pinsker's inequality

$\delta(P_n, P) \leq \sqrt{\frac{1}{2} KL(P_n || P)} \to 0$

$\delta(P, P_n) \leq \sqrt{\frac{1}{2} KL(P || P_n)} \to 0$

And $\delta(P_n, P) = \delta(P, P_n)$.

**Appendix F (Additional Experiments)**
**Comparison of convergence speed in different loss functions**

As the Figure5 below, we trained our DDSP ConNet in different loss functions in the learning rate of 0.001 and it shows the convergence speed of these different loss functions. It is illustrated that the convergence speed of DSC loss, Jaccard loss and re-weighted cross entropy were approximately similar, but DSC loss and Jaccard loss were far more stable than re-weighted cross entropy and remained the comparably same scores in different training data, where DSC loss performed best at the beginning of training. In conclusion, DSC loss and Jaccard loss are more appropriate in training network and these experiments validated the conclusion of topology strength of them.

Figure5: Convergence speed of different loss functions when training our DDSP network.

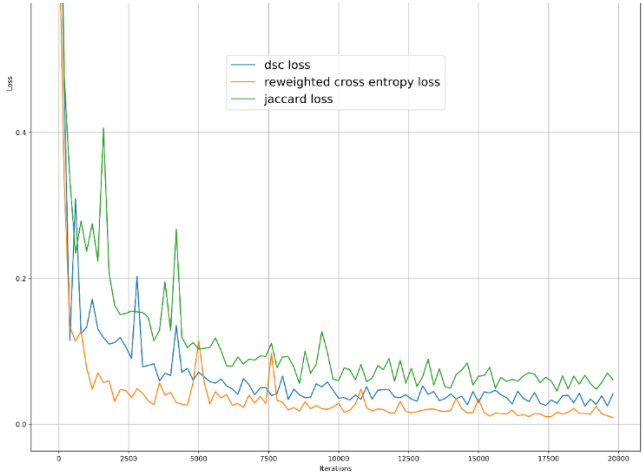

## Ablation Study for Residual Long Connection

Long connection methods may influence the performance of the segmented model. Therefore, in order to compare the performance of different long connection methods and find their influences on the whole network, we performed several control experiments with different long connection methods. In these experiments, we adopted the structure of our DDSP with DSC loss, and all the other parameters remained the same.

Table4. Cross validation performance of our model on Dice coefficient with different long connection methods.

| methods | Dice coefficient [%] |
|---|---|
| **Without long connection** | 86.13 |
| **Residual long connection** | 86.42 |
| **Concatenation long connection** | 84.96 |

It is illustrated in the Table4 that the residual long connection outperformed the other two ways. On the other words, element-wisely concatenation is more beneficial to preserve multi-level context and achieve richer information, so the residual long connection helps improve the accuracy of prostate segmentation in our DDSP ConNet.

## Ablation Study for Multi-Supervision

As we all know, multi-supervision method will effectively enhance CNN models performance, but we do not know its qualitative detail, so we did experiments in our DDSP ConNet with diverse supervise loss weights. Firstly we set the total loss weights to one, and then gave the different weights. The baseline is our DSPP model with residual long connection in DSC loss function.

Table5. Performance of our model on Dice coefficient with different loss weights configurations. The configurations from left to right are the output of master branch, stage_2 loss branch and stage_3 loss branch.

| Loss weights configuration | Dice coefficient [%] |
|---|---|
| [1, 0, 0] (Baseline) | 84.35 |
| [0.5, 0.333, 0.167] | 85.20 |
| [0.6, 0.25, 0.15] | 85.50 |
| [0.8, 0.15, 0.05] | 86.42 |
| [0.8, 0.2, 0] | 85.90 |
| [0.9, 0.075, 0.025] | 86.01 |

The loss weight configuration in 0.8-0.15-0.05 yielded the best performance. Compare to the baseline, it outperformed with an improvement of 2.07 in terms of Dice coefficient. Additionally, we could see that, to some extent, the larger loss weight of master branch yielded better performance. For instance, compare the configuration [0.8, 0.15, 0.05] with [0.8, 0.2, 0], the former achieved better segmented results, so the auxiliary supervision of stage_3 is beneficial for the parameters to converge into optimum and thus obtain more precise segmented model.

**Ablation Study for Multi-Fusion Output**

As mentioned before, we adopt the multi-supervision, and there are multi branches in our model, where each branch outputs segmentation results in different precision. In order to highly reuse these information, in validation and testing stages, we adopt multi-fusion output. Here we did experiments on different output ways to validate the effectiveness of multi-fusion output. In these experiments, we adopted our DDSP ConNet and all the other parameters remained the same.

Table6. Performance of our model on Dice coefficient with different multi-fusion output configurations. The configurations from left to right are the output of master branch, stage_2 loss branch and stage_3 loss branch, where 1 represents using this branch's output, 0 represents non-use.

| Multi-Fusion output | Dice coefficient [%] |
|---|---|
| [1, 0, 0] (without) | 86.66 |
| [1, 1, 1] | 86.98 |

Table6 clearly illustrates that, compared to only using the master branch, multi-fusion output could improve the performance of segmented model. In other words, the branches in stage_2 and stage_3 preserve useful contexture information and also produce useful segmented masks.